\documentclass{article}

\usepackage[dblblindworkshop, final]{neurips_2025}

\usepackage[utf8]{inputenc} 
\usepackage[T1]{fontenc}    
\usepackage{hyperref}       
\usepackage{url}            
\usepackage{booktabs}       
\usepackage{amsfonts}       
\usepackage{nicefrac}       
\usepackage{microtype}      
\usepackage{xcolor}         
\usepackage{graphicx}
\usepackage{subcaption}

\title{Explainable AI-Generated Image Detection RewardBench}
\workshoptitle{The First Workshop on Generative and Protective AI for Content Creation}

\author{%
  \begin{tabular}{@{}c@{}}
  Michael Yang\textsuperscript{1} \quad
  Shijian Deng\textsuperscript{1} \quad
  William T.~Doan\textsuperscript{1} \\
  Kai Wang\textsuperscript{2} \quad 
  Tianyu Yang\textsuperscript{3} \quad
  Harsh Singh\textsuperscript{4} \quad
  Yapeng Tian\textsuperscript{1}
  \end{tabular}
  \\[4pt]
  \textsuperscript{1}The University of Texas at Dallas \quad
  \textsuperscript{2}University of Toronto \\
  \textsuperscript{3}University of Notre Dame \quad
  \textsuperscript{4}Stony Brook University \\[2pt]
  \texttt{ymichael469@gmail.com;} \quad
  \texttt{\{shijian.deng, doan, yapeng.tian\}@utdallas.edu} \\
  \texttt{kaikai.wang@mail.utoronto.ca} \quad
  \texttt{tyang4@nd.edu} \quad
  \texttt{singh17@cs.stonybrook.edu}
}

\begin{document}

\maketitle

\begin{abstract}
    Conventional, classification-based AI-generated image detection methods cannot explain why an image is considered real or AI-generated in a way a human expert would, which reduces the trustworthiness and persuasiveness of these detection tools for real-world applications. Leveraging Multimodal Large Language Models (MLLMs) has recently become a trending solution to this issue. Further, to evaluate the quality of generated explanations, a common approach is to adopt an "MLLM as a judge" methodology to evaluate explanations generated by other MLLMs. However, how well those MLLMs perform when judging explanations for AI-generated image detection generated by themselves or other MLLMs has not been well studied. We therefore propose \textbf{XAIGID-RewardBench}, the first benchmark designed to evaluate the ability of current MLLMs to judge the quality of explanations about whether an image is real or AI-generated. The benchmark consists of approximately 3,000 annotated triplets sourced from various image generation models and MLLMs as policy models (detectors) to assess the capabilities of current MLLMs as reward models (judges). Our results show that the current best reward model scored 88.76\% on this benchmark (while human inter-annotator agreement reaches 98.30\%), demonstrating that a visible gap remains between the reasoning abilities of today's MLLMs and human-level performance. In addition, we provide an analysis of common pitfalls that these models frequently encounter. Code and benchmark are available at~\url{https://github.com/RewardBench/XAIGID-RewardBench}.

\end{abstract}

\begin{figure}
    \centering
    \includegraphics[width=0.8\linewidth]{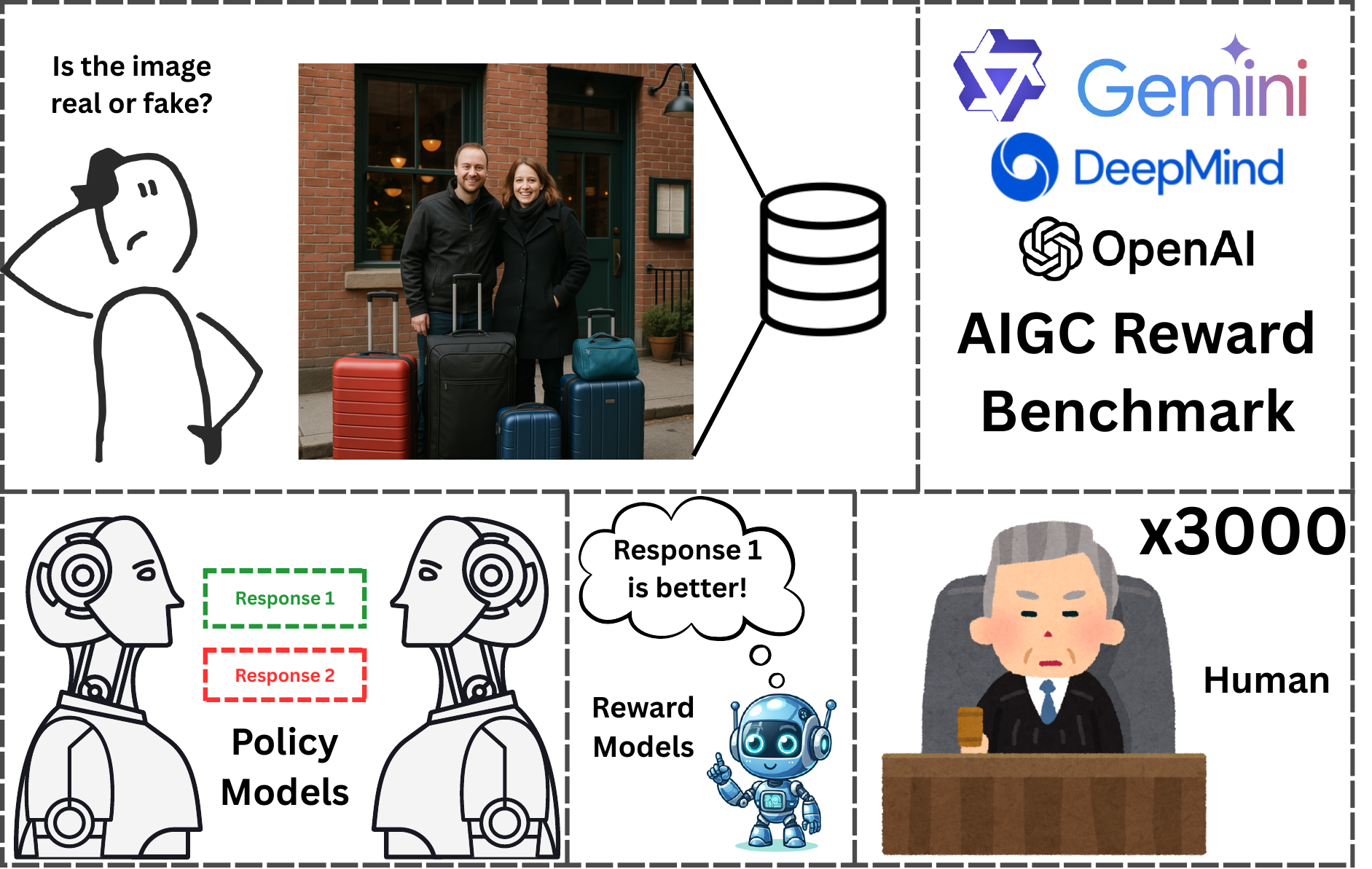}
    \caption{Overview of the XAIGID-RewardBench. With the capabilities of today's MLLMs, an AI-generated image detection response can include both a verdict and an explanation, providing much more insight into the reasoning process. Furthermore, MLLMs are also now being used to judge the quality of these explanations, but little work has been done to assess the quality of the MLLM judges themselves. Therefore, we present a human-annotated benchmark to evaluate how well these reward models (i.e., MLLM judges) score the explanations provided by MLLMs for AI-generated image detection.}
    \label{fig:teaser}
\end{figure}

\section{Introduction}

AI-generated and modified images are becoming increasingly prevalent, causing widespread damage to information ecosystems and public trust. Thus, the development of autonomous systems to detect such images is crucial. However, current methods typically provide a classification decision without the kind of explanation or justification that a human expert can offer, which limits the persuasiveness and trustworthiness of their outputs. To address this, an ideal model should not only perform binary classification but also analyze the image comprehensively, leveraging logic, fine-grained details, real-world knowledge, and even tool use to determine authenticity and articulate its reasoning with supporting evidence.

Multimodal Large Language Models (MLLMs) offer a promising approach to generating such nuanced explanations. By processing visual details, an MLLM can investigate an image's authenticity through a deductive process, much like a detective examining a crime scene. However, these explanations generated by models are not always accurate; challenges such as hallucinations and faulty reasoning remain significant obstacles. To mitigate these issues in a scalable way, a reward model "judge" can be implemented to evaluate the validity of a generated explanation, either during training or inference.

This raises a critical research question: \emph{how is the judge itself validated?} If such a judge is to be reliable, its own performance must be rigorously benchmarked. To that end, we present the \textbf{Explainable AI-Generated Image Detection (XAIGID) RewardBench}. This benchmark contains approximately 3,000 triplets, each consisting of an image and two competing explanations of its authenticity (Figure~\ref{fig:teaser}). It is designed to evaluate reward models (i.e., "judges") on their ability to determine the better response. These annotations were carefully curated and manually verified, adhering to a strict rubric to minimize bias. This benchmark can help identify which MLLMs are most suited to act as reward models for judging explanations from AI-generated image detectors.

Our primary contributions are as follows:
\begin{enumerate}
    \item We are the first to conduct a systematic study of current MLLMs as both policy models and reward models within the context of explainable AI-generated image detection.
    \item We introduce the first reward model benchmark designed to gauge MLLMs' performance on the novel task: Judging Explainable AI-Generated Image Detection.
    \item We provide meaningful insights into the capabilities and common failure modes in explainable AI-generated image detection.
\end{enumerate}

\section{Related Works}
\label{sec:related}

\subsection{MLLMs for Explainable AIGC Detection}
Early AI-generated content (AIGC) detectors emphasized binary decisions with limited interpretability. Recent work has turned to MLLMs to couple detection with explanation and, in some cases, grounding. LEGION~\cite{kang2025legion} introduces SynthScars, pixel-level artifact masks with textual rationales, and an MLLM pipeline that localizes artifacts and verbalizes why they indicate synthesis. FakeXplained~\cite{ji2025interpretable} fine-tunes MLLMs to align rationales with evidence via detection, box-level grounding, and coherent textual explanations. MLLM-DEFAKE~\cite{ji2025towards} benchmarks off-the-shelf MLLMs against dedicated detectors and humans, proposing a reasoning-driven pipeline for explainable decisions. ForenX~\cite{tan2025forenx} adds a "forensic prompt" and an explanation resource (ForgReason) to steer MLLMs toward human-aligned forensic cues. FakeShield~\cite{xufakeshield} further formalizes explainable image forgery detection and localization for MLLMs, coupling authenticity decisions with pixel-level masks and textual justifications. In parallel, RADAR~\cite{wang2025radar} targets semantic fakes that violate world knowledge, pairing bounding boxes with textual descriptions.
For the video modality, ExDDV~\cite{hondru2025exddv} provides 5.4K real/deepfake clips with human text rationales and click supervision, showing that both textual and spatial supervision benefit explainable VLMs. Complementing these advances, Forensics-Bench~\cite{wang2025forensics} offers a comprehensive LVLM evaluation suite spanning multiple modalities and tasks (binary decisions, localization, attribution), highlighting the performance gap that remains for current models. Orthogonally, LVLM-DFD~\cite{yuunlocking} improves generalization and explanation via a Knowledge-guided Forgery Detector and a Forgery Prompt Learner that align forensic patterns with LVLM reasoning. Altogether, these works establish that MLLMs can produce human-readable justifications and sometimes explicit evidence (boxes/masks); however, they only evaluate policy models, not the reliability of reward models (judges) for evaluating explainable AIGC detection.

\subsection{Reward Models and Benchmarks}
Reinforcement learning has recently been explored for MLLMs in AIGC detection, but primarily with rule-based reward functions. Veritas~\cite{tan2025veritas} trains an MLLM detector with Mixed Preference Optimization and pattern-aware GRPO to shape planning/self-reflection during reasoning about forgeries, improving OOD generalization on the HydraFake dataset. FakeXplained~\cite{ji2025interpretable} uses GRPO-style rewards on verdict correctness, grounding quality, and rationale format after SFT. For judging explanations, previous work has either relied on direct human judgment or used a popular MLLM as a judge, without systematically evaluating the MLLM judges themselves with a standardized reward-model benchmark in the AI-generated image domain.
General reward-model benchmarks exist: RewardBench~\cite{lambert2025rewardbench} and its update RewardBench 2~\cite{malik2025rewardbench} assess textual reward models across instruction following, reasoning, and safety; VL-RewardBench~\cite{li2025vl} and Multimodal RewardBench~\cite{yasunaga2025multimodal} extend this evaluation to VLM judges across perception, hallucination detection, reasoning, knowledge, and safety. These resources demonstrate broad trends for reward models but do not probe forensic explanation quality or artifact-grounded reasoning on AI-generated images.
To our knowledge, no prior work provides a domain-specific reward-model benchmark for judging explainable AI-generated image detection. XAIGID-RewardBench fills this gap by supplying human-verified, 4-way-annotated triplets (image, two explanations) that directly measure how well MLLMs act as judges for explainable AI-generated image detection.

\section{XAIGID-RewardBench}
\label{sec:XAIGID-RewardBench}

We first document the entire creation process of our XAIGID-RewardBench, including image dataset synthesis, response generation with MLLMs as AI-generated image detectors, and RewardBench annotation. Then we formalize the novel task: Judging Explainable AI-Generated Image Detection.

\subsection{Image Dataset}

For real images, we used COCO-2017~\cite{lin2014microsoft}. We assume "real" images do not change dramatically over time, so using a classic dataset is sufficient for providing non-AI-generated images. We sampled 500 random images from COCO-2017 and then cropped them to a square resolution.

To cover a diverse range of AI-generated images, we selected 10 representative image generation models to generate 500 images as the other half of our image dataset. We selected 10 representative image generation models, listed below, emphasizing recent SOTA models while also including several classic ones.

\paragraph{Image Generation Models.}
The generation models used in this study include Stable Diffusion 1.5~\cite{Rombach_2022_CVPR}, Stable Diffusion 3.5~\cite{esser2024scaling}, Imagen 3~\cite{baldridge2024imagen}, Imagen 4~\cite{imagen4_model_card_2025}, Glide~\cite{nichol2022glide}, GPT-Image-1~\cite{openai_4o_image_generation_2025}, Recraft V3~\cite{recraft_design_language_2024}, Flux.1 Schnell, Flux.1 Dev~\cite{flux2024}, and Bagel \cite{deng2025bagel}.

We generated 500 images in total, 50 per model. To ensure the topic distribution of generated images was similar to that of the real images, we used Qwen 2.5 3B to generate basic, unbiased, and detailed captions of the real images. From there, we partitioned the 500 captions into ten groups of 50. Each group of captions was then used to generate 50 images with one of the ten generation models. This approach allows for pairing real images with generated images derived from similar descriptions, making the detection task more diverse and challenging.

\subsection{Policy Models' Detection Response with Explanation}

We utilized several state-of-the-art MLLMs as AI-generated image detectors to generate responses. In total, we employed seven different models to provide a diverse pool of answers: Qwen-VL 2.5 3B, Qwen-VL 2.5 7B~\cite{Qwen2.5-VL}, Gemini 2.5 Flash, Gemini 2.5 Pro~\cite{comanici2025gemini}, OpenAI o3~\cite{openai_o3_o4_mini_2025}, GPT-4o~\cite{hurst2024gpt}, and Gemma-3n-4B~\cite{gemma_3n_2025}.

Each model was given the image and asked to determine the image's authenticity, followed by an explanation. We used the following prompt:

\textit{Is the following image real or fake in terms of AI generation? Give a single-word answer ('real' or 'fake') to start your response. Then, provide a more detailed explanation.}\footnote{Due to API safety policies, some responses from o3 and GPT-4o were refused, but the vast majority of responses yielded long, rich answers, all at least one paragraph long and often multiple paragraphs.}

\begin{figure}
    \centering
    \includegraphics[width=0.9\linewidth]{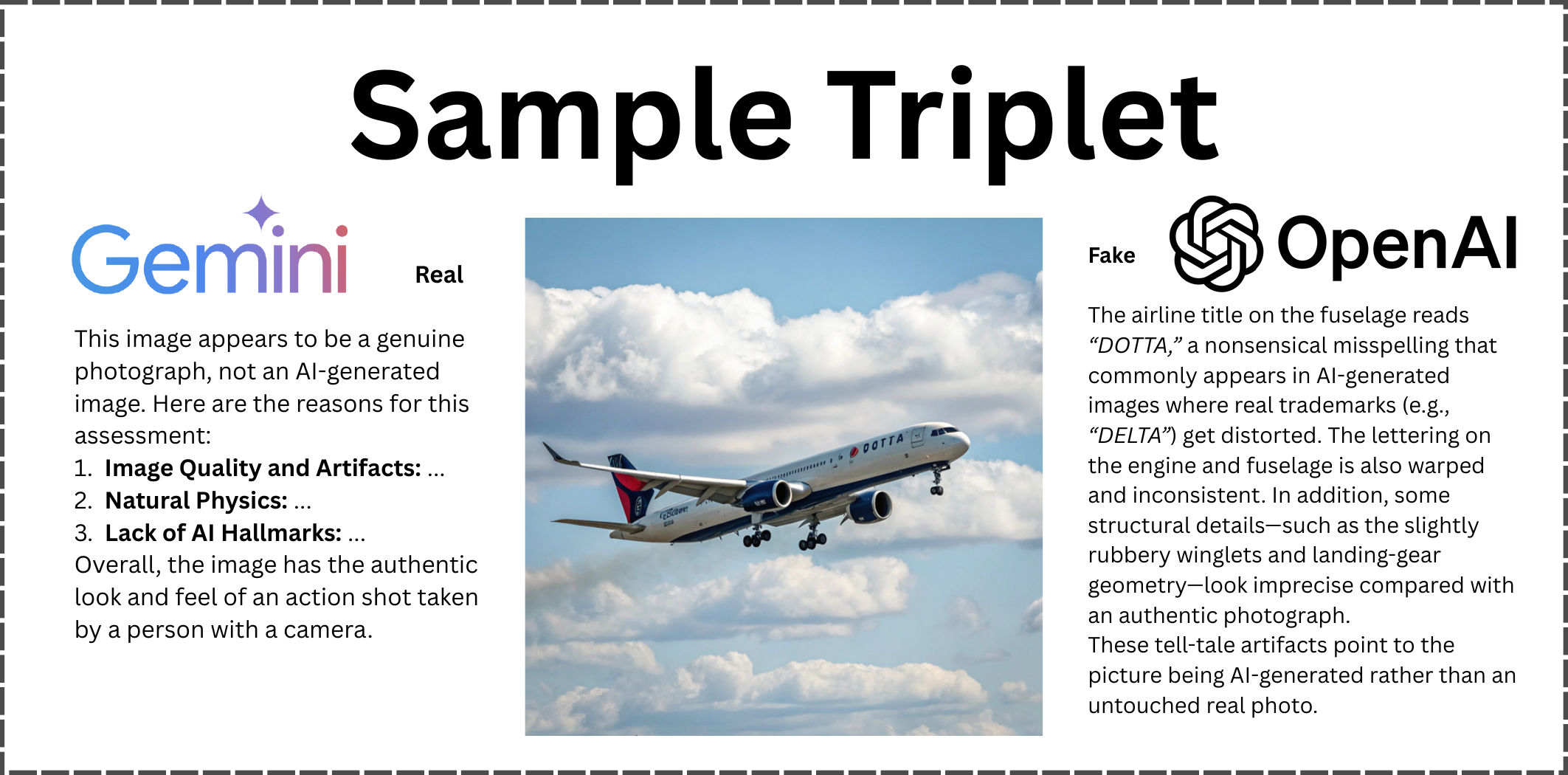}
    \caption{A visual depiction of the input (a triplet) for MLLM judges. Both responses come from SOTA MLLMs as AI-generated image detectors, yet they disagree with each other. An MLLM judge must deduce which one is more convincing given all 3 pieces of information (One image and two detection responses from detectors). MLLM judges currently still cannot achieve human performance on the stated task.}
    \label{fig:sampletriplet}
\end{figure}

\paragraph{Results.}

We calculated the accuracy of these models against our dataset to gauge their raw competence at detecting AI-generated images (Table~\ref{tab:overall_acc_table},~\ref{tab:class_acc_table_sources}).

\begin{table}[h]
    \caption{Accuracy of MLLMs as policy models (AI-generated image detectors). Performance on detecting AI-generated images is high among proprietary models, and both o3 and GPT-4o exhibit high accuracy and capability on this task. A stronger MLLM generally performs better on AI-generated image detection tasks. However, Qwen 2.5 3B outperforms its 7B counterpart; we hypothesize this is because the 7B model's increased capability leads to overconfidence. While the 3B model correctly identifies real images, its performance on fake images is close to random chance. Conversely, the 7B model exhibits more complex reasoning, but this often leads it to be confidently incorrect.}
    \label{tab:overall_acc_table}
    \centering
    \resizebox{0.7\textwidth}{!}{
        \begin{tabular}{lrrr}
            \toprule
            \textbf{Policy Model} & \textbf{Overall Acc} & \textbf{Real Image Acc} & \textbf{Fake Image Acc} \\
            \midrule
            \multicolumn{4}{l}{\textit{Proprietary}} \\
            GPT-4o & 92.28\% & 94.18\% & 90.38\% \\ 
            o3 & 88.87\% & 99.80\% & 77.96\% \\ 
            Gemini-2.5 Pro & 78.90\% & 99.60\% & 58.20\% \\ 
            Gemini-2.5 Flash & 75.48\% & 98.60\% & 52.40\% \\
            \midrule
             \multicolumn{4}{l}{\textit{Open Source}} \\
            Qwen 2.5 3B & 71.70\% & 96.80\% & 46.60\% \\ 
            Gemma 3n E4B-it & 56.80\% & 98.20\% & 15.40\% \\ 
            Qwen 2.5 7B & 52.10\% & 98.60\% & 5.60\% \\
            \bottomrule
        \end{tabular}
    }
\end{table}

\begin{table}[h]
    \caption{Policy model accuracy on detecting images from various generation models. The accuracies show that images from the Imagen 3 and Imagen 4 models were the most difficult to detect.}
    \label{tab:class_acc_table_sources}
    \scriptsize 
    \centering
    \resizebox{1\textwidth}{!}{
        \begin{tabular}{lrrrrrrrrrr}
            \toprule
            \textbf{Policy Model} & \textbf{SD 1.5} & \textbf{Imagen 3} & \textbf{Bagel} & \textbf{Dev} & \textbf{Imagen 4} & \textbf{Recraft} & \textbf{Schnell} & \textbf{SD 3.5} & \textbf{GPT-Image-1} & \textbf{Glide} \\
            \midrule
            \multicolumn{11}{l}{\textit{Proprietary}} \\
            GPT-4o & 100.00\% & 68.00\% & 98.00\% & 96.00\% & 94.00\% & 56.00\% & 98.00\% & 98.00\% & 96.00\% & 100.00\% \\ 
            o3 & 100.00\% & 40.82\% & 96.00\% & 90.00\% & 72.00\% & 20.00\% & 92.00\% & 90.00\% & 78.00\% & 100.00\% \\ 
            Gemini-2.5Pro & 98.00\% & 24.00\% & 88.00\% & 70.00\% & 24.00\% & 12.00\% & 54.00\% & 62.00\% & 66.00\% & 84.00\% \\ 
            Gemini-2.5 Flash & 92.00\% & 18.00\% & 88.00\% & 62.00\% & 18.00\% & 12.00\% & 36.00\% & 46.00\% & 58.00\% & 94.00\% \\ 
            \midrule
            \multicolumn{11}{l}{\textit{Open Source}} \\
            Qwen 2.5 3B & 32.00\% & 28.00\% & 80.00\% & 64.00\% & 56.00\% & 2.00\% & 46.00\% & 38.00\% & 54.00\% & 66.00\% \\ 
            Gemma 3n E4B-it & 16.00\% & 14.00\% & 24.00\% & 8.00\% & 14.00\% & 0.00\% & 8.00\% & 4.00\% & 22.00\% & 44.00\% \\ 
            Qwen 2.5 7B & 10.00\% & 0.00\% & 14.00\% & 4.00\% & 2.00\% & 0.00\% & 4.00\% & 2.00\% & 4.00\% & 16.00\% \\ 
            \bottomrule
        \end{tabular}
    }
\end{table}

\paragraph{Pairing.}
We then created 3 triplets per image for a total of 3,000 initial triplets. Pairing was conducted randomly in a round-robin fashion to ensure equal distribution. The triplets were then scrambled and sent for annotation (Figure~\ref{fig:sampletriplet}).

\subsection{RewardBench Annotation}

Objectively deciding which response is better can be challenging for annotators. They may be affected by many personal biases, even subtle hidden ones. To combat this, we established a general rubric derived from our own qualitative analysis of MLLM responses. More details are provided in Section~\ref{sec:additional}.

\paragraph{Annotation.}

To perform annotation, annotators were given the target image and the two responses. They were also presented with the original prompt to understand what the model was responding to. Lastly, they were allowed to make one of four selections to annotate:
\textit{
\begin{itemize}
    \item "First": To denote that the first response was superior.
    \item "Second": To denote that the second response was superior.
    \item "Tie": Both responses were equally good in quality and not completely bad.
    \item "Both Bad": Both responses provided unsatisfactory explanations.
\end{itemize}
}

\paragraph{Filtering.}
When multiple annotators label the same triplet, conflicts can arise. Conflicts can be resolved by majority vote. If ties are encountered, they can be resolved manually. Additionally, any annotations where too little time was spent (e.g., annotations completed in an unusually short time) can also be filtered out.

In a side experiment, we also evaluated which models generated the most favorable responses. The table below reveals the models' performance, ranked using an Elo system. Each model started at 1200 Elo to begin (Table~\ref{tab:elo_rankings}). \footnote{Ties were excluded when calculating Elo ratings.}

\begin{table}[h]
    \caption{Elo ratings of policy models (based on human judges). Both Gemini models showed high performance as policy models, while the open-source models lagged behind in the Elo ranking. The longer, richer responses, most of the time, were much better than the short, brief, and shallow responses from the open-source policy models.}
    \label{tab:elo_rankings}
    \centering
    \resizebox{0.7\textwidth}{!}{
        \begin{tabular}{rlrrrrr}
            \toprule
            \textbf{Rank} & \textbf{Policy Model} & \textbf{Elo Rating} & \textbf{Games} & \textbf{Wins} & \textbf{Losses} & \textbf{Win Rate} \\
            \midrule
            1 & Gemini 2.5 Pro & 1836.0 & 859 & 810 & 49 & 0.943 \\
            2 & Gemini 2.5 Flash & 1702.2 & 846 & 758 & 88 & 0.896 \\
            3 & o3 & 1261.5 & 753 & 397 & 356 & 0.527 \\
            4 & GPT-4o & 941.1 & 731 & 242 & 489 & 0.331 \\
            5 & Qwen 2.5 3B & 929.3 & 694 & 145 & 549 & 0.209 \\
            6 & Qwen 2.5 7B & 902.0 & 684 & 150 & 534 & 0.219 \\
            7 & Gemma 3n E4B & 827.9 & 689 & 126 & 563 & 0.183 \\
            \bottomrule
        \end{tabular}
    }
\end{table}

\begin{figure}
    \centering
    \includegraphics[width=0.3\linewidth]{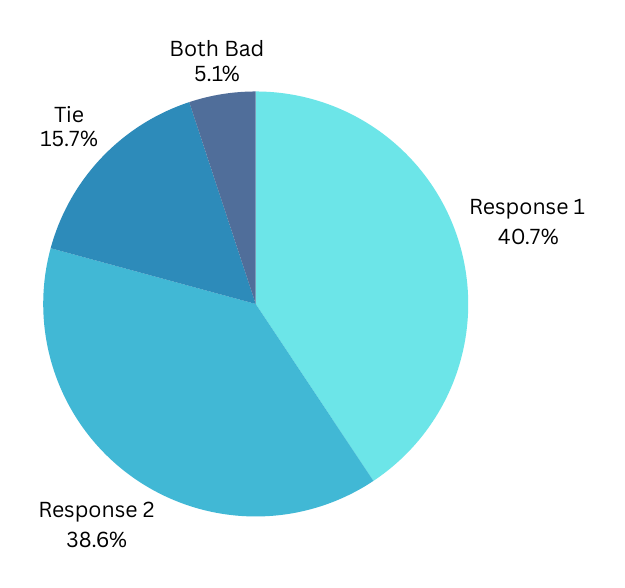}
    \caption{Distribution of annotator choices. While Response 1 was picked 2.1\% more often, the difference is not significant. This indicates that there was no significant position bias in the human annotation.}
    \label{fig:choicedistributionpiechart}
\end{figure}

We also analyzed the choice distribution to check for potential annotation bias, such as choosing Response 1 solely because it appeared first. Our data shows that our annotations were unbiased by the relative positions of the responses (Figure~\ref{fig:choicedistributionpiechart}).

\paragraph{Inter-human agreement.}

Disparities in human preferences are natural. To maximize the effectiveness of the rubric, we conducted an experiment where two annotators were given a previously unseen batch and asked to annotate all 100 triplets. After doing so, we compared their annotations to evaluate the rubric's effectiveness in aligning human preferences.

Our results show that the human annotators agreed on 68.0\% of the annotations and 98.3\% on annotations where both annotators decided on a clear winner (considering only annotations without "Tie" or "Both Bad").

Overall, these statistics show that our rubric was effective at aligning the annotations in a vast majority of cases. The reason for the large disparity between 4-way agreement and clear-winner agreement could be that annotators have differing thresholds for "high-quality" vs. "low-quality." In a case where both responses in a triplet may be "low-quality," one annotator might mark the triplet as "Both Bad," while another might decide that one is slightly better than the other. Regardless, a 98.3\% agreement for annotations with a hard preference shows that annotators were still able to agree in most cases on which response was better.

\subsection{Task: Judging Explainable AI-Generated Image Detection}
\label{sec:task}

The goal of the task is to evaluate how well MLLMs, acting as reward models, can judge explanations generated by MLLMs as detectors in AI-generated image detection.

Let $\mathcal{G}$ denote a set of image generation models. Each $G\in\mathcal{G}$ induces a distribution over images given a caption $c$ and randomness $z$, written $I = G(c,z)$. Let $\mathcal{I}_{\mathrm{real}}$ be a set of photographs sampled from a real-image distribution, and let $\mathcal{I}_{\mathrm{gen}} = \{\,G(c,z): G\in\mathcal{G}\,\}$ be generated images. We define $\mathcal{I}=\mathcal{I}_{\mathrm{real}}\cup\mathcal{I}_{\mathrm{gen}}$ and a ground-truth authenticity label $l:\mathcal{I}\to\{0,1\}$ where $l(I)=1$ for real and $l(I)=0$ for AI-generated.

A policy model (MLLM detector) $\pi\in\Pi$, given a policy prompt $p_p$ maps an image to a structured response
\[
\ r \;=\; \pi(I, p_p) \;=\; (y,e), \qquad y\in\{0,1\},\; e\in\mathcal{E},
\]
where $y$ is the one-word verdict and $e$ is a free-form explanation. Detector (policy) accuracy is
\[
\mathrm{Acc}_{\mathrm{det}}(\pi) \;=\; \frac{1}{|\mathcal{I}|}\,\big|\{\,I\in\mathcal{I}:\; y(\pi(I, p_p)) = l(I)\,\}\big|.
\]

For reward modeling (judging), a triplet $t$ consists of a single image and two competing policy responses to the same prompt $p_p$: $t=(I,r_a,r_b)$ with $r_a=(y_a,e_a)$ and $r_b=(y_b,e_b)$. A judge (MLLM as a reward model) $J\in\mathcal{J}$ prompted with $p_j$: \emph{Between the two responses, the original prompt, and the image, which response is better?} outputs a 4-way preference
\[
J(t,p_p,p_j)\in\mathcal{Y}, \qquad \mathcal{Y}=\{\mathrm{First},\mathrm{Second},\mathrm{Tie},\mathrm{BothBad}\},
\]
indicating whether the first response is better, the second is better, both are comparably good, or both are unsatisfactory.

A human judge $H$ provides labels in the same space $\mathcal{Y}$ and produces the ground-truth preference \[H(t,p_p,p_j)\in\mathcal{Y}.\]

Let $\mathcal{T}$ be the set of annotated triplets with ground-truth labels $H_t$. The judge's 4-way accuracy is
\[
\mathrm{Acc}^{(4)}(J) \;=\; \frac{1}{|\mathcal{T}|}\,\big|\{\,t\in\mathcal{T}:\; J_t=H_t\,\}\big|.
\]
For a 2-way accuracy on the subset $\mathcal{T}_2=\{\,t\in\mathcal{T}:\; H_t\in\{\mathrm{First},\mathrm{Second}\}\,\}$:
\[
\mathrm{Acc}^{(2)}(J) \;=\; \frac{1}{|\mathcal{T}_2|}\,\big|\{\,t\in\mathcal{T}_2:\; J_t=H_t\,\}\big|.
\]

\section{Experiment and Analysis}

We performed experiments based on our newly created benchmark. This allowed us to evaluate how current state-of-the-art MLLMs perform as judges when evaluating explainable AI-generated image detection.

\subsection{Experimental Setup}
We reused all seven policy models that we used to generate initial responses as judges. They were utilized as described in Section~\ref{sec:task}. Each model was presented with all annotated triplets, asked to respond with their preference among the four options followed by an elaboration, and then scored based on their alignment with human preferences.

\subsection{Results}

To evaluate each model's performance as a judge, we calculated the accuracy it achieved when compared with human annotations. The results are presented in Table~\ref{tab:judgeacc}.

\begin{table}[h]
    \caption{Performance of MLLMs as Reward Models (Judges). All proprietary models and Gemma performed reasonably well on this benchmark. Gemma's strong performance may be related to its origin as a model from the Gemini family. The two Qwen models are much weaker and thus performed much worse on the benchmark. The '2-Way Acc' column considers only cases where the human annotator marked a clear winner. The goal was to see if models could identify the correct winner in unambiguous cases. This is because ties create "gray areas" which make it harder to classify, but the task is more straightforward when a clear winner exists.}
    \label{tab:judgeacc}
    \centering
    \resizebox{0.8\textwidth}{!}{
        \begin{tabular}{lrr}
            \toprule
            \textbf{Reward Model (MLLM as a Judge)} & \textbf{4-Way Acc (\%)} & \textbf{2-Way Acc (\%)} \\
            \midrule
            \multicolumn{3}{l}{\textit{Proprietary}} \\
            Gemini 2.5 Pro & 68.35 & \textbf{88.76} \\
            Gemini 2.5 Flash & 68.92 & 87.26 \\
            GPT-4o & 60.62 & 85.59 \\
            o3 & 64.74 & 82.17 \\
            \midrule
            \multicolumn{3}{l}{\textit{Open Source}} \\
            Gemma 3n e4b & 66.88 & 84.54 \\
            Qwen 2.5 7B & 29.37 & 82.28 \\
            Qwen 2.5 3B & 31.35 & 77.12 \\
            \bottomrule
        \end{tabular}
    }
\end{table}

These accuracies shine a light on the ability of today's SOTA models to act as reward models for AI-generated images. While these models are able to identify high-quality responses to an extent, their ability to perform this judge role still has room for improvement. No model achieved an accuracy over 90\% on this benchmark, which reveals the gap between human and current MLLM judges (Table~\ref{tab:judgeacc}).

\subsection{Analysis}
Beyond just analyzing qualitative results, we found multiple factors that can play a role in an explanation's quality. This is crucial for the development of both policy and reward models, and these findings can also provide better insight on how to construct a more thorough rubric or standard for judging explainable AI-generated image detection.

\begin{figure}[h]
    \centering
    \begin{subfigure}{0.48\linewidth}
        \centering
        \includegraphics[width=\linewidth]{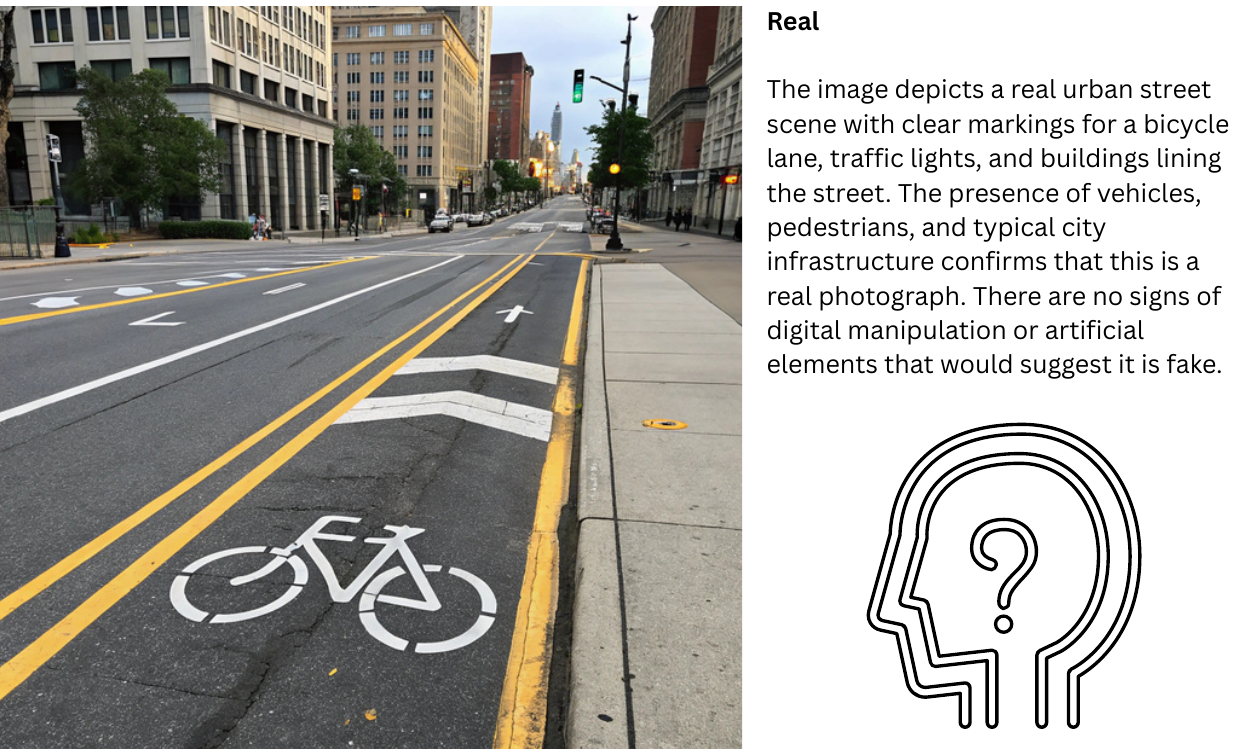}
        \caption{Example of a response making an irrelevant point. While the mention of the city and the road is true, there is no actual argument being made. An AI-generated image can also have vehicles and pedestrians.}
        \label{fig:uselesspoints}
    \end{subfigure}
    \hfill
    \begin{subfigure}{0.48\linewidth}
        \centering
        \includegraphics[width=\linewidth]{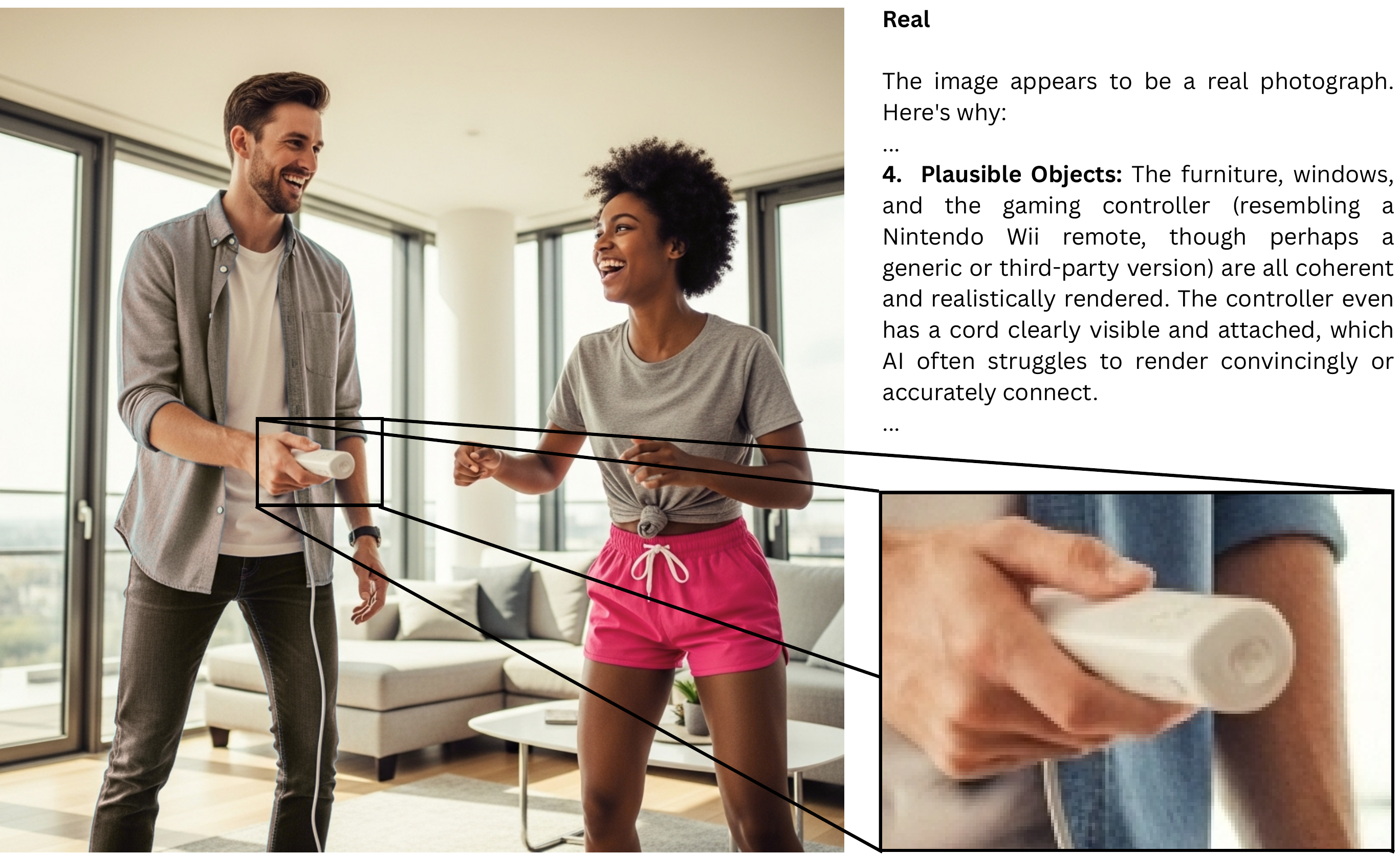}
        \caption{Example of a model generalizing details and missing key artifacts. While the controller's shape looks realistic, the model overlooks its details, failing to notice the missing buttons.}
        \label{fig:generalizingdetail}
    \end{subfigure}
    \caption{Some failure cases of MLLM policy models.}
    \label{fig:combined_figures}
\end{figure}

\paragraph{Even one mistake should qualify an image as AI-generated.}
Frequently, a model correctly identifies anomalies within an image (Figure~\ref{fig:onemistakeisenough}). However, the model assumes that because the rest of the image looks fine, the image is probably real and that the anomaly was just a "hiccup." In reality, authentic images do not have such "hiccups." It only takes one piece of evidence to show that an image is fake.

\paragraph{Simply stating a detail doesn't provide any convincing argument.}
Models occasionally state something true about the image (Figure~\ref{fig:uselesspoints}). While their statement about the image is true, it does not prove anything. Something needs to be stated about the mentioned detail, or why the existence of that detail proves that an image is real or fake. This shortcoming highlights the gap between the ability of current SOTA models to identify and then reason.

\paragraph{Models tend to generalize details in media.}
Instead of looking into an area that might cause suspicion, a model might just skip over it and say it "looks realistic" (Figure~\ref{fig:generalizingdetail}). By doing so, they miss the key anomaly that proves why an image is fake. Humans still hold the ability to look at an image, identify which areas might be suspicious, and then further scrutinize that region for clues. A current known issue is that models commonly hallucinate details. A more pertinent issue is that models are not identifying everything that they should be.

\paragraph{Longer responses are not equivalent to better answers.}
During annotation, we instructed annotators to look beyond just length and pursue responses that held real, meaningful arguments. For example, Gemini is known to present larger, more detailed arguments, and while these responses are usually better, this is not always the case. Hallucination, meaningless points (Figure~\ref{fig:sampletriplet}), and poor analysis of details can sometimes occur in long responses. In fact, a shorter response that correctly spots all the anomalies is better. A shorter response that accomplishes more is preferable to a long response that accomplishes the same level of reasoning.

\subsection{Ablation Study}

\paragraph{Modern real-world image generalization.}

As stated earlier, our real image dataset is compiled from COCO 2017. The motivation behind this decision was that it was a readily available dataset with a large variety of "real" images. However, a concern was that the dataset was outdated. Our assumption was that real images do not change that much over time. To test this hypothesis, we took 100 real-world images ourselves with a variety of backgrounds, scenes, and angles. We then had the models respond to the same original prompt and compared the accuracy they achieved on guessing the image's authenticity.

The average accuracy for COCO 2017 images was 98.99\%, while the accuracy for our new images was 98.29\%. While COCO 2017 was slightly easier, the difference is negligible. This small margin shows that COCO 2017 is a sufficient real-image dataset.

\paragraph{Human against policy models.}

In addition to our Elo calculations among MLLMs as policy models, we wanted to see how a human response would fare against MLLMs in providing explanations in detecting AI-generated images. We had annotators write 100 new human responses. We then paired each human response with responses from three other random AI models to create a new batch of 300 triplets. After conducting the human judging process again, we provided the new Elo results in Table~\ref{tab:elo_rankings_human}.

\begin{table}[h]
    \caption{Elo ratings of policy models including a human baseline. As shown, the human is able to compete with other SOTA MLLMs in response generation but surprisingly loses to both Gemini models. This may be attributed to the fact that Gemini can generate more comprehensive responses while human-generated responses are shorter and with fewer details. This indicates that MLLMs can be more practical as policy models (not as reward models yet) than humans to generate explanations in AI-generated image detection, as they are good at generating detailed and comprehensive explanations which are more appealing to human judges.}
    \label{tab:elo_rankings_human}
    \centering
    \resizebox{0.7\textwidth}{!}{
        \begin{tabular}{rlrrrrr}
            \toprule
            \textbf{Rank} & \textbf{Policy Model} & \textbf{Elo Rating} & \textbf{Games} & \textbf{Wins} & \textbf{Losses} & \textbf{Win Rate} \\
            \midrule
            \multicolumn{7}{l}{\textit{Proprietary Models}} \\
            1 & Gemini 2.5 Pro & 1474.7 & 36 & 34 & 2 & 0.944 \\
            2 & Gemini 2.5 Flash & 1386.2 & 38 & 31 & 7 & 0.816 \\
            4 & o3 & 1218.8 & 16 & 9 & 7 & 0.562 \\
            5 & GPT-4o & 1191.9 & 26 & 12 & 14 & 0.462 \\
            \midrule
            \multicolumn{7}{l}{\textit{Open Source Models}} \\
            6 & Qwen 2.5 3B & 1083.0 & 19 & 4 & 15 & 0.211 \\
            7 & Qwen 2.5 7B & 1002.5 & 27 & 3 & 24 & 0.111 \\
            8 & Gemma 3n & 1002.1 & 26 & 3 & 23 & 0.115 \\
            \midrule
            \multicolumn{7}{l}{\textit{Human Reference}} \\
            3 & Human & 1240.9 & 188 & 92 & 96 & 0.489 \\
            \bottomrule
        \end{tabular}
    }
\end{table}

\section{Conclusion}

In this paper, we presented the first systematic study of MLLMs as both policy (AI-generated image detectors) and especially reward models (judges) for the tasks related to explainable AI-generated image detection, with an emphasis on judging explanations for AI-generated image detection.
Our work delivers three key contributions. First and foremost, we introduce \textbf{XAIGID-RewardBench}, the first benchmark designed to evaluate MLLM reward models (judges) in this novel task: Judging Explainable AI-Generated Image Detection, composed of approximately 3,000 human-annotated preference triplets. Second, through this benchmark, we quantitatively demonstrate a visible gap in current model capabilities, with our top-performing reward model reaching only $88.76\%$ accuracy (while human inter-annotator agreement reaches $98.30\%$). Third, we provide meaningful insights into common failure modes, identifying critical flaws in model reasoning such as generalizing over clear artifacts and failing to connect evidence to a logical conclusion.
Collectively, our contributions provide a foundational benchmark and a clear analysis to guide the community in developing more accurate, reliable, and trustworthy systems for detecting and explaining AI-generated images, including judging explainable AI-generated image detection.

\bibliographystyle{plain}
\bibliography{custom}

\newpage
\appendix

\section{Additional Information for Benchmark Annotation}
\label{sec:additional}

\paragraph{Rubric:}
\textit{
\begin{itemize}
    \item Hallucination: Any extraneous details?
    \item Details: Did it get all the important details(distortions)?
    \item Logic: Did it correctly formulate an argument(regardless of wrong/right)?
    \item Relevance: Do the points that the LLM make actually help its argument?
    \item Counterargument: Did it shut down any opposing points?
    \item Weighing: Did it correctly weigh why the image might be real or fake(Lots of points may be thrown out, but there needs to be a tying thread to unify all the points in favor of its argument)
    \item Contradiction: Does the model contradict itself?
    \item Insight: Are the arguments shallow or unrefined?
\end{itemize}
}
\section{Setup of Annotation Tool}

\begin{figure}[ht]
    \centering
    \includegraphics[width=0.6\linewidth]{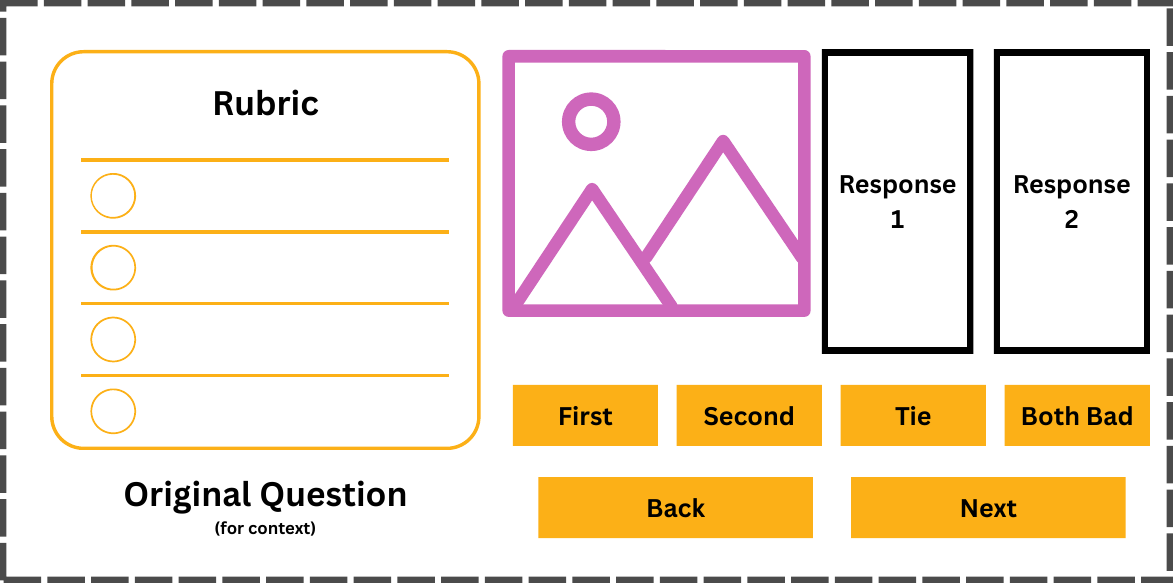}
    \caption{A conceptual illustration of our annotation tool. The image and response sources are hidden to remove bias. Annotators are also encouraged to look past the initial one-word verdict and focus more on the explanation given by MLLM detectors.}
    \label{fig:annotationsneakpeak}
\end{figure}

Human annotators might be prone to bias, such as choosing longer answers simply because of their length or judging a response purely based on its final answer rather than its thought process. Our goal is to maximize the MLLM's useful reasoning steps and minimize nonsensical output. To ensure minimal bias in annotation, several countermeasures were taken. First, a hidden stopwatch recorded the time spent on each annotation as metadata. If the recorded time was under fifteen seconds, the annotation was filtered out. Second, cropping the images helps hide the image's true nature. Some annotators may be inclined to select the response that agrees with their own initial assessment. Thus, cropping the images helps encourage annotators to truly think about the model's response (Figure~\ref{fig:annotationsneakpeak}).

\section{AI-Generated Image Examples}

\begin{figure}[ht]
    \centering
    \includegraphics[width=0.8\linewidth]{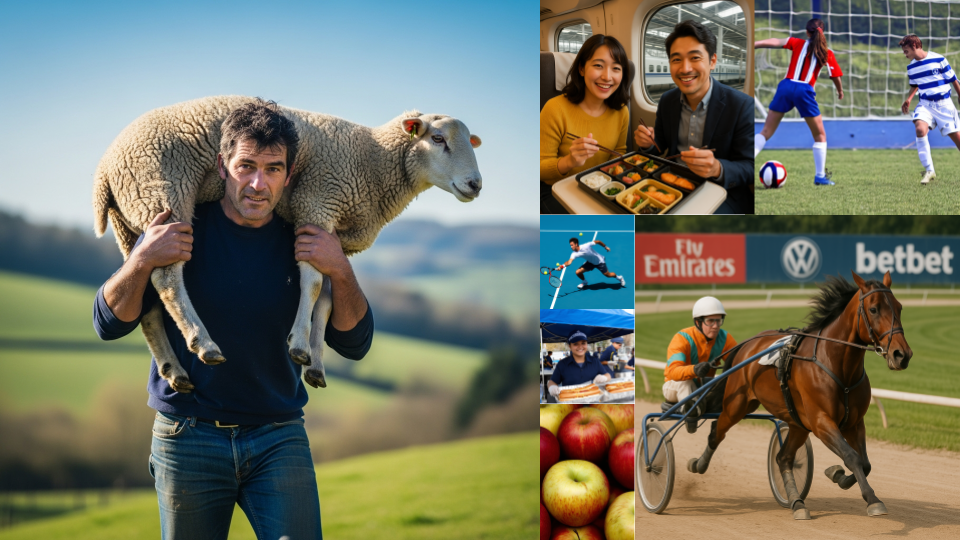}
    \caption{A collage of several AI-generated images from our benchmark. Some of these images were challenging to identify, even for human eyes. }
    \label{fig:fakeimagecollage}
\end{figure}

We show some examples of generated images from our benchmark in Figure~\ref{fig:fakeimagecollage}.

\section{More Response Example}

\begin{figure}
    \centering
    \includegraphics[width=1\linewidth]{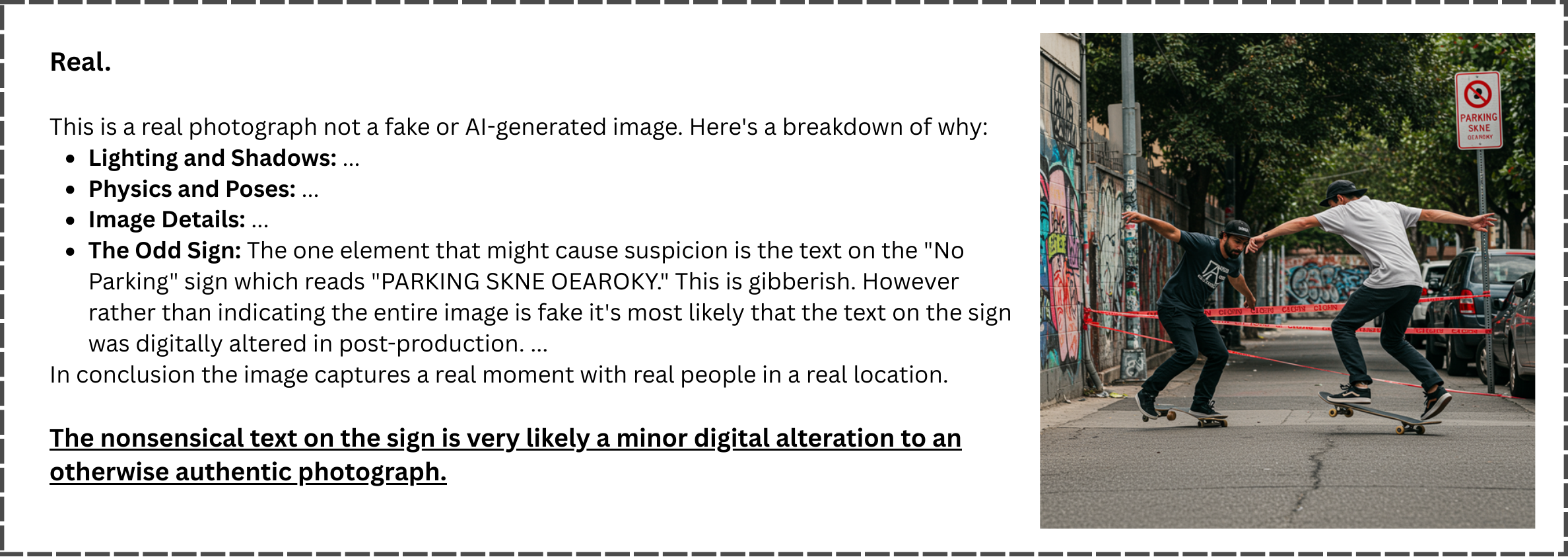}
    \caption{Example of a model downplaying a critical artifact. The distorted text on the street sign is a dead giveaway that the image is generated. While the rest of the image is very consistent, a mistake like this is undeniable. Still, the model tries to argue that one slip-up is acceptable.}
    \label{fig:onemistakeisenough}
\end{figure}

We provide additional examples of policy model responses in Figure~\ref{fig:onemistakeisenough},~\ref{fig:uselesspoints}, and~\ref{fig:generalizingdetail}.

\section{Limitations and Future Work}

While this benchmark provides a crucial tool for the community, our reliance on human preference as the "ground truth" for explanation quality has its own limitations. However, at present, no absolutely objective metric exists for evaluating explanations in AI-generated image detection, considering the complex nature of the task itself. Therefore, human judgment remains the only viable gold standard. We did our best to make this process rigorous by using a detailed rubric, and our high inter-annotator agreement confirms a consistent signal. Future work, however, could achieve even greater robustness by dramatically increasing the scale of the dataset and employing a larger, more diverse pool of annotators to build a stronger consensus on challenging or ambiguous cases.

Similarly, our initial focus on static images was a pragmatic choice to ensure depth and quality before tackling more complex data types. A key direction for future work is to build upon this foundation to create benchmarks for more modalities, such as AI-generated videos, synthesized audio, and even generated 3D assets, each presenting unique forensic challenges. This expansion is critical for developing comprehensive and versatile detection systems.

Ultimately, the goal is to detect AI-generated content and explain the detection process in a trustworthy way. The path forward lies in moving beyond using MLLMs on their own and toward building MLLM-powered agents that can actively reason and leverage external tools. These agents could dynamically use forensic tools and construct evidence-based narratives, mirroring a human expert's workflow. The central challenge will be to develop reward models that can guide this complex reasoning and action process, a necessary step to close the gap between current models and the dynamic, trustworthy analysis required for real-world applications.

\end{document}